\documentclass[journal]{IEEEtran}
\usepackage{algorithm, algorithmic}
\usepackage{svg}
\usepackage{booktabs}
\usepackage{placeins}
\usepackage{longtable}
\usepackage{times}
\usepackage{amsmath}
\usepackage{latexsym}

\usepackage{amssymb} 
\usepackage{cleveref}
\usepackage{siunitx}
\usepackage{multirow}
\usepackage{svg}
\ifCLASSINFOpdf
\else
\fi

\begin{document}
%

\title{Weak-Link Optimization for Multi-Agent Reasoning and Collaboration}
%
%
%
\author{
    \IEEEauthorblockN{
        Haoyu Bian$^{1}$,
        Chaoning Zhang$^{1}$,
        Jiaquan Zhang$^{1}$,
        Xingyao Li$^{1}$,
        Yuanfang Guo$^{2}$,
        Wei Dong$^{3}$,
        Yang Yang$^{1}$,\\
    }
    \IEEEauthorblockA{$^{1}$ University of Electronic Science and Technology of China, Chengdu 611731, China}\\
    \IEEEauthorblockA{$^{2}$ Beihang University, Beijing 100191, China}\\
    \IEEEauthorblockA{$^{3}$ Xi'an University of Architecture and Technology, Xi’an  710064, China}\\

    \IEEEauthorblockA{Email:chaoningzhang@uestc.edu.cn}\\
    
\thanks{This work was supported by the National Natural Science Foundation of China (NSFC) under the General Program (Grant No. 62572104).}
}

\maketitle

\begin{abstract}
LLM-driven multi-agent frameworks address complex reasoning tasks through multi-role collaboration. However, existing approaches often suffer from reasoning instability, where individual agent errors are amplified through collaboration, undermining overall performance. Current research mainly focuses on enhancing high-capability agents or suppressing unreliable outputs to improve framework effectiveness, while systematic identification and reinforcement of performance-limiting agents receive less attention. To address this gap, we propose WORC, a \underline{w}eak-link \underline{o}ptimization framework for multi-agent \underline{r}easoning and \underline{c}ollaboration, grounded in the weak-link principle. WORC follows a two-stage workflow. In the weak agent localization stage, task features are constructed, and a meta-learning-based weight predictor trained on optimal configurations identified by swarm intelligence algorithms (SIAs) enables zero-shot mapping from these features to agent performance weights, where the agent with the lowest predicted weight is identified as the weak agent. In the weak-link optimization stage, an uncertainty-driven allocation strategy assigns additional reasoning budgets to weak agents, with lower predicted weights leading to larger repeated-sampling quotas to compensate for reliability deficiencies. Experimental results show that WORC achieves an average accuracy of 82.2\% on reasoning benchmarks while improving framework stability and cross-architecture generalization, suggesting that compensating for weak links, rather than reinforcing strengths alone, enhances the robustness of multi-agent systems.

\end{abstract}


%
\IEEEpeerreviewmaketitle

\section{Introduction}

\IEEEPARstart{L}{arge} Language Models (LLMs) have demonstrated remarkable capabilities in generative natural language processing tasks \cite{chang2024survey,he2025intelligent}, yet they continue to underperform in mathematical problem-solving and logical reasoning. In response, researchers proposed reasoning methods such as Chain of Thought (CoT) \cite{wei2022chain}, which formalize human reasoning approaches into prompt templates and emphasize subtask decomposition and multi-step reasoning. Recent studies have further explored task-driven alignment and structure-aware reasoning-chain optimization \cite{zhang2026text,zhang2026learning}. Concurrently, the emergence of AI Agents \cite{masterman2024landscape}, particularly multi-agent frameworks \cite{talebirad2023multi} leveraging planning, reflection, and tool utilization capabilities across collaborating specialized agents, has significantly enhanced LLMs' performance on complex problem-solving tasks \cite{wang2024rethinking}. Recent advances further extend collaborative reasoning beyond static agent cooperation toward interaction-aware \cite{zhang2025social}, role-adaptive \cite{liu2025rcr}, and consensus-driven collaboration paradigms \cite{yang2026dynamic}, enabling more structured deliberation processes across distributed reasoning agents. These developments improve the effectiveness of multi-agent reasoning systems in long-horizon decision-making tasks, including scientific programming \cite{ren2025towards}, medical diagnosis \cite{zuo2025kg4diagnosis}, and autonomous planning \cite{wu2025multi,li2026experience}.

\begin{figure}[t]
  \centering
  \includegraphics[width=\linewidth]{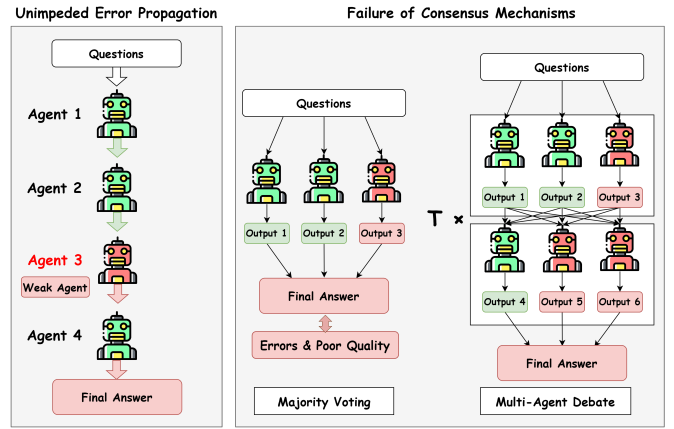}
  \caption{Overview of the vulnerability of weak agents in multi-agent reasoning. Sequential reasoning propagates errors from underperforming agents. Majority voting and multi-agent debate mitigate but do not eliminate the influence of weak agents.}
  \label{fig:weak}
\end{figure}

However, in complex reasoning tasks, multi-agent frameworks face substantial coordination challenges, requiring task decomposition and collaboration to align individual agent actions with overall objectives \cite{li2024survey}. The reliability of a reasoning path depends on the compounded reliability of its components, making the system inherently susceptible to performance degradation \cite{zheng2025rethinking}. Individual underperforming agents in a multi-agent architecture, hereafter referred to as \textbf{weak agents}, compromise the overall reliability of the system by inducing inaccurate reasoning, unreliable decisions, and error-prone outputs. Conventional design paradigms, which emphasize stronger reasoning agents or incorporate simple consensus mechanisms such as voting \cite{fu2025deep} and debate \cite{liang2024encouraging}, remain susceptible to instability and exhibit high performance variability despite their effectiveness \cite{choi2025debate}. This fragility manifests specifically as:
\begin{enumerate}
\item \textbf{Error accumulation across reasoning stages}: In task decomposition, outputs of preceding agents serve as inputs for subsequent ones. Low-accuracy or miscalibrated outputs from any agent may propagate errors downstream, amplifying their impact.
\item \textbf{Consensus degradation under heterogeneous agent reliability}: Consensus mechanisms rely on agreement among agents. Erroneous contributions from limited-capability agents may degrade overall decision quality and introduce systemic biases.
\end{enumerate}
Similarly, multi-path reasoning approaches alleviate these risks by exploring multiple candidate reasoning trajectories; however, their effectiveness remains constrained by weak agents and may introduce additional computational overhead \cite{chen2024comm}.



To address these limitations, we propose WORC, a reasoning optimization framework for LLM-driven multi-agent systems grounded in the weak-link optimization principle. This perspective is inspired by the bottleneck-driven system optimization principle, commonly referred to as the ``Buckets Effect,'' which has been widely adopted in system reliability engineering, production optimization, and fault-tolerant distributed system design, where overall system performance is constrained by its weakest components. In the context of multi-agent reasoning, this motivates a shift toward targeted compensation of weak agents to enhance reasoning reliability.


To operationalize this principle, WORC adopts a two-stage optimization process consisting of weak agent localization and weak-link optimization. In the weak agent localization stage, SIAs are employed to estimate optimal agent weight vector configurations based on multi-agent reasoning performance over sampled task-type datasets, thereby capturing task-dependent agent contributions within collaborative reasoning processes. This formulation leverages the population-based global search capability of swarm intelligence methods to model agent performance distributions without requiring explicit supervision, and constructs the resulting weight vectors as a knowledge base for cross-task generalization. When new reasoning tasks are encountered, task signatures are constructed using text embedding models such as OpenAI embeddings, incorporating semantic mean embeddings and structural statistical features. These signatures are subsequently processed by a meta-learning-based weight predictor to retrieve the most relevant weight vector from the knowledge base as a benchmark for weak agent identification. In the weak-link optimization stage, an automatic budget allocation mechanism assigns additional reasoning resources to the identified weak agents based on the predicted weight configuration. All agents then generate candidate solutions according to their allocated quotas, and the final output is obtained through a voting-based aggregation module.

As a demonstration of this generalizable optimization method, we design a simple chain-based multi-agent reasoning system called AgentChain (AC) as an illustrative implementation of our approach. Additionally, we conduct comprehensive evaluations across different datasets and tasks, demonstrating the framework's enhanced reasoning capabilities, stability, and interpretability. Our contributions include:
\begin{enumerate}
 \item We propose an optimization method for LLM-driven multi-agent reasoning, inspired by the ``weak-link'' principle, which focuses on enhancing system robustness by addressing weak components in the architecture.
 \item To generalize weak agent detection across different tasks, we construct a meta-learning weight predictor and SIAs for task feature analysis, enabling zero-shot identification of weak agents across tasks.
 \item Comprehensive experimental evaluations and theoretical analyses demonstrate the method's effectiveness in enhancing reasoning accuracy and system stability across various multi-agent frameworks.
\end{enumerate}

\section{Related Work}

\subsection{Multi-Agent Systems for Reasoning}
Multi-Agent Systems represent one of the key architectural paradigms for large language models (LLMs) \cite{han2024llm,talebirad2023multi}, enabling distributed reasoning through collaboration and interaction among multiple agents \cite{ke2025survey}. Multi-agent architectures lead to improved reasoning performance by enabling structured decomposition of reasoning processes through collaborative interaction \cite{hong2023metagpt}. The application of multi-agent frameworks addresses several limitations found in single-agent reasoning. Chen et al. \cite{chen2024comm} demonstrate that by constructing a multi-agent, multi-reasoning path framework, where language models play different roles and collaborate, task-solving performance on complex scientific problems can be improved. Similarly, Gu et al. \cite{gu2025explain} showed that breaking complex tasks into subtasks and employing pipeline-style multi-agent collaboration facilitates large models in tackling complex reasoning problems.

Existing multi-agent reasoning enhancement methods exhibit several notable limitations:
\begin{itemize}
    \item \textbf{Majority voting} treats all agents equally, failing to identify or downweight weak or unreliable agents \cite{wang2024survey}.
    \item \textbf{Self-consistency} mechanisms reinforce correlated erroneous reasoning trajectories by repeatedly emphasizing similar inference paths \cite{wu2025personalized}.
    \item \textbf{Debate-based approaches} are susceptible to destabilization when incorrect or misleading arguments dominate the discussion process \cite{liu2024groupdebate}.
    \item \textbf{Static weight allocation} ignores task-specific and context-dependent variations in agent performance \cite{putta2024agent}.
\end{itemize}



\subsection{Meta-Learning and Task Adaptation}
The core objective of meta-learning is to enable models to leverage prior task distributions to rapidly adapt to new tasks \cite{hospedales2021meta}. Early research such as MAML \cite{finn2017model} optimized parameter initialization for fast task adaptation across tasks. Subsequently, methods like Prototypical Networks \cite{snell2017prototypical} demonstrate efficient few-shot inference via task-level representations in few-shot classification scenarios. With the rise of LLMs, meta-learning concepts have been applied to enhance LLMs' generalization and adaptation capabilities \cite{sinha2024maml}.

Introducing meta-learning into multi-agent reasoning systems aims primarily to improve agents' collective collaboration abilities, enabling overall joint adaptation dynamics. Current research mainly follows two paths. The first involves building agents with meta-cognitive abilities (such as ReMA \cite{wan2025rema}, MetaMind \cite{zhang2025metamind}) that enable planning, monitoring, and adjustment of their own reasoning processes. The second focuses on meta-level coordination \cite{sun2025multi}, directly optimizing collaboration patterns between agents through game theory or meta-learning strategies, allowing systems to quickly form efficient team reasoning structures. However, this field still faces significant challenges, including high computational costs and excessive dependence on foundation model capabilities \cite{bilal2025meta}.

\subsection{Swarm Intelligence Algorithms in LLMs}
Swarm Intelligence Algorithms (SIAs) have been widely studied as effective tools for solving complex optimization problems \cite{chakraborty2017swarm} by mimicking collective behaviors observed in biological systems. Classical algorithms, including Particle Swarm Optimization (PSO) \cite{kennedy1995particle} and Grey Wolf Optimizer (GWO)\cite{mirjalili2014grey}, established population-based stochastic optimization frameworks for continuous and combinatorial optimization problems. Recently, novel SIAs have continued to emerge with refined search mechanisms and enhanced optimization performance. For example, the Marine Predator Algorithm \cite{faramarzi2020marine} simulates different hunting behaviors based on the encounter rate between prey and predators. The Hippopotamus Optimization (HO) Algorithm \cite{amiri2024hippopotamus} simulates hippos' territorial marking and defensive attack behaviors.

Recent studies have brought SIAs into LLMs and deep learning architectures, showing clear improvements in optimization efficiency, parameter tuning, and handling of complex reasoning tasks \cite{tang2021review}. For example, researchers use SIAs' parallel search abilities to explore parameter and collaboration policy search spaces in neural reasoning systems \cite{kouziokas2023swarm}. However, while SIAs have been widely applied in traditional multi-agent computer systems, their application in LLM-driven multi-agent systems remains underexplored, with examples like SwarmSys \cite{li2025swarmsys} and AMRO-S \cite{wang2026efficient} introducing SIAs to achieve scalable and adaptive reasoning. Nevertheless, this integration still contains significant gaps in research.

\begin{figure*}[h]
  \centering
  \includegraphics[width=\linewidth]{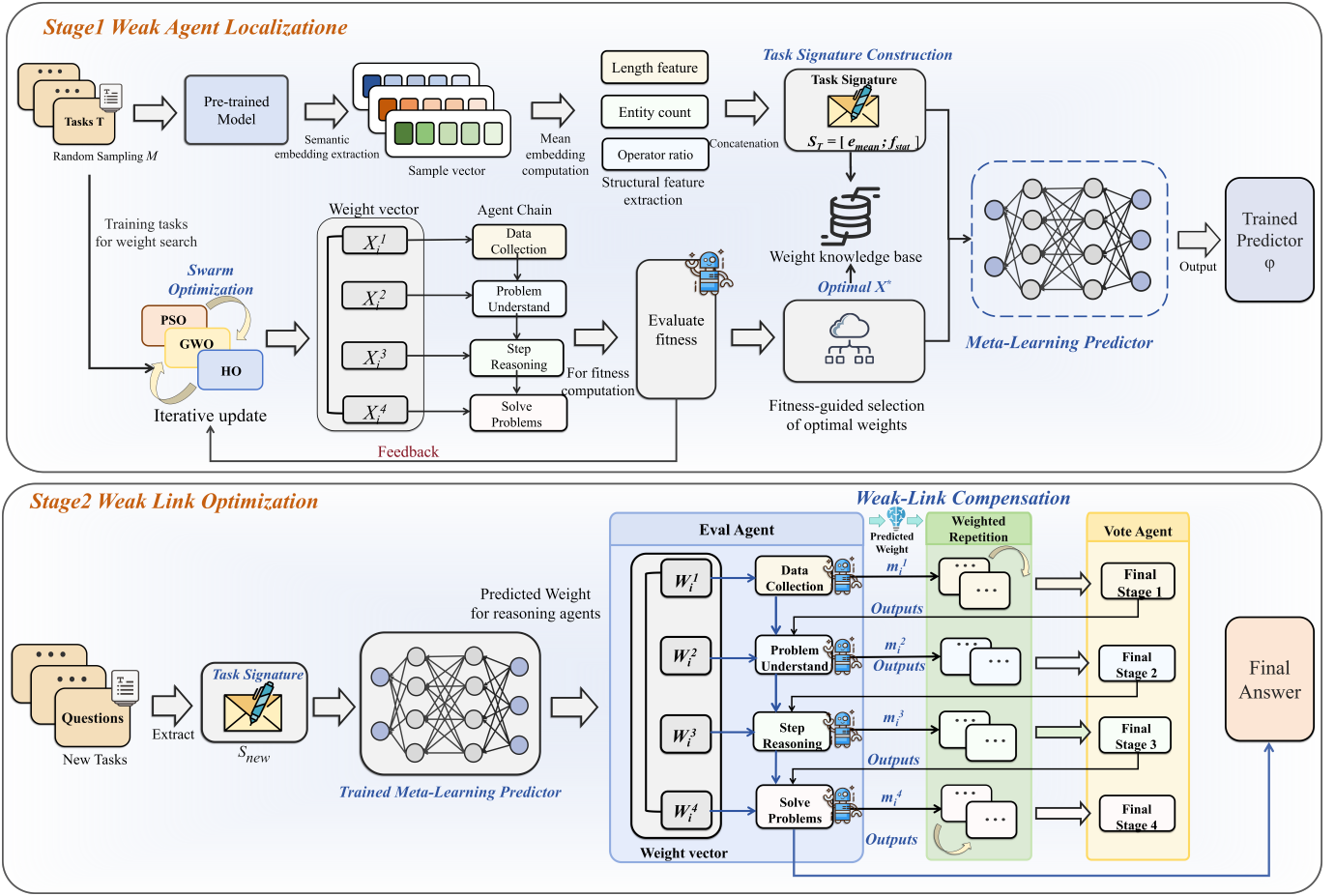}
  \caption{Overview of the WORC method in the AC framework.  
(a) Weak Agent Localization: A weight knowledge base is constructed via SIA training, and task signatures are generated. A meta-learning predictor outputs the most suitable weight vector for a new task, enabling identification and assessment of weak agents.  
(b) Weak Agent Optimization: The predicted weight vector guides targeted allocation of reasoning budget to compensate low-performing agents. Agents execute sequentially within the AgentChain framework, with VoteAgent selecting the best outputs to achieve collaborative reasoning and performance optimization.
}
  \label{fig:first}
\end{figure*}

\section{Method}
\label{sec:method}
This section presents WORC demonstrated through implementation on the AC framework, illustrating the process of weak agent localization and targeted optimization across multiple tasks, as shown in Figure \ref{fig:first}.

\subsection{Weak Agent Localization}
To ensure effective weak agent identification while maintaining generalization, this stage comprises three components: weight knowledge base construction, task signature development, and meta-learning weight predictor design.

\subsubsection{Weight Knowledge Base Construction}
\label{sec:3.1.1}
The weight knowledge base is constructed through SIAs training, serving as the foundation for method generalization. This base stores instructive weight vector collections that numerically model individual agent performance within multi-agent frameworks across different tasks under optimal scoring conditions.

SIAs and multi-agent frameworks operate in distributed paradigms where individual entities achieve global optimal collaboration through local interactions, with dynamic individual contribution assessment mechanisms providing natural modeling foundations for multi-agent weight configuration \cite{feng2024model}. In this application scenario, SIAs primarily reveal individual agent performance during multi-agent reasoning processes and model them numerically as weight vectors. Consequently, we simplify and redefine the SIAs' core designs for method adaptation. During initialization, we define an initial population containing K candidate solution groups (answer sets for few-shot scenarios) as weight configuration schemes for multi-agent collaborative frameworks:

\begin{equation}
\begin{split}
\mathcal{P}^0 = \{ X_1^0, X_2^0, \ldots, X_K^0 \}.
\end{split}
\end{equation}

For a single candidate solution group $i$ in training iteration $t$ with few-shot learning:

\begin{equation}
\begin{split}
X_i^t = (x_{i,1}^t, x_{i,2}^t, \ldots, x_{i,N}^t),
\end{split}
\end{equation}
where $(x_{i,1}^t, x_{i,2}^t, \ldots, x_{i,N}^t)$ represents the weight distribution among different agents, where each dimension corresponds to an agent's weight, and $N$ represents the number of agents.

The optimization process follows an iterative method, with each iteration starting from position vector $X_{curr}^t$, corresponding to the highest accuracy known in the current training. In each iteration, position updates follow this formula:

\begin{equation}
\begin{split}
X_i^{t+1} = X_{curr}^t + \Delta X_i^t,
\end{split}
\end{equation}
where position change $\Delta X_i^t$ integrates multiple information sources:

\begin{equation}
\begin{split}
\Delta X_i^t = C_1 \cdot (P_{curr,i} - X_i^t) + C_2 \cdot (G_{curr} - X_i^t) \\+ C_3 \cdot (X_{random} - X_i^t).
\end{split}
\end{equation}

In the above equation, $P_{curr,i}$ is the local optimal solution found by individual $i$. $G_{curr}$ is the global optimal solution found by the entire group. $X_{random}$ is a randomly generated position to promote exploration. $V_i^t$ is the movement trend of individual $i$ at current iteration $t$. $C_1$, $C_2$, and $C_3$ are parameters that may have randomness or variation with iterations. After each iteration, we apply two constraints: binding weights to $[0.05, 1.0]$ and normalizing them to sum to 1 to ensure effective subsequent calculations.

Notably, the binding between weight vector $X^t = (x_1^t, x_2^t, \ldots, x_N^t)$ and agent reasoning effectiveness stems from the objective function of the optimization process. Let $f(X)$ represent the accuracy function of the system on the training set; the optimization problem can be formalized as:
\begin{equation}
\begin{split}
\max_X f(X) \text{ s.t. } \sum_{j=1}^N x_j = 1, x_j \in [0.05, 1.0] \forall j .
\end{split}
\end{equation}

At the optimal solution $X^*$, weight value $x_j^*$ reflects agent $j$'s marginal contribution $\frac{\partial f}{\partial x_j}(X^*)$. The essence of this binding mechanism is that the optimization algorithm aligns weight allocation with agents' marginal contributions through iterative updates. If agent $j$'s reasoning output contributes significantly to improving overall accuracy, increasing $x_j$ will significantly enhance $f(X)$, resulting in a larger $x_j^*$ in the optimal solution. However, this mapping cannot be implemented directly and requires an external evaluation mechanism, with specific evaluation methods explained in Section \ref{sec:3.2.2}.

After few-shot learning on this type of dataset, the weights with the best overall accuracy performance among candidate solution groups will be used as the best weights for building the knowledge base. In practice, the construction of the weight knowledge base is performed over a collection of few-shot samples drawn from multiple related reasoning task datasets. The corresponding optimal weight vectors obtained from these sampled tasks are aggregated to form the knowledge base as a set of task-representative weight configurations. This design ensures that generalizable guidance data is obtained using only a small number of dataset samples and limited computational resources. The specific design and theoretical analysis of SIAs are detailed in the supplementary materials.

\subsubsection{Task Signature}
\label{sec:3.1.2}
The construction of task signatures is designed to enable weak agent identification across tasks, providing a dense, continuous semantic similarity measurement foundation for subsequent meta-learning weight predictors. When a multi-agent architecture encounters a new reasoning task, WORC constructs a task signature $\mathbf{s}_T$ for each task $T$, which integrates features from both semantic and structural dimensions. Specifically, first, $M$ unlabeled samples $\{x_j\}_{j=1}^M$ are randomly drawn from a reasoning task dataset, where $M$ is identical to the few-shot sample size used for SIAs weight optimization in Section \ref{sec:3.1.1}, ensuring that task signature construction and weight vector generation are derived from the same sampled task instances. And a pretrained embedding model (such as OpenAI Embedding \cite{xian2024vector}) is used to obtain semantic representations, with the mean embedding calculated as the semantic index of the task:
\begin{equation}
\mathbf{e}_j = \text{Embed}(x_j) \in \mathbb{R}^d, \quad \mathbf{e}_{\text{mean}} = \frac{1}{M}\sum_{j=1}^M \mathbf{e}_j.
\end{equation}

Further, four statistical measures are extracted to capture the structural characteristics of the task:
\begin{enumerate}
\item Length features: mean and variance of sample lengths
\begin{equation}
f_{\text{len}}^{(1)} = \frac{1}{M}\sum_{j=1}^M |x_j|, \quad f_{\text{len}}^{(2)} = \sqrt{\frac{1}{M}\sum_{j=1}^M (|x_j| - f_{\text{len}}^{(1)})^2}
\end{equation}
\item Entity features: average entity count
\begin{equation}
f_{\text{ent}} = \frac{1}{M}\sum_{j=1}^M \text{EntityCount}(x_j)
\end{equation}
\item Operator features: proportion of logical and arithmetic symbols
\begin{equation}
f_{\text{op}} = \frac{1}{M}\sum_{j=1}^M \frac{\text{Count}(\{+, -, \times, \div, \text{if}, \text{because}\}, x_j)}{|x_j|}
\end{equation}
\end{enumerate}

These features are concatenated into a statistical vector $\mathbf{f}_{\text{stat}} = [f_{\text{len}}^{(1)}, f_{\text{len}}^{(2)}, f_{\text{ent}}, f_{\text{op}}]$, which is then combined with the semantic embedding to form the final task signature:

\begin{equation}
\mathbf{s}_T = [\mathbf{e}_{\text{mean}}; \mathbf{f}_{\text{stat}}] \in \mathbb{R}^{d+k}
\end{equation}

These task signatures are stored in the knowledge base alongside their corresponding weight vectors. Through this unified embedding space, task type similarity can be quantitatively characterized via distances between task signature vectors \cite{achille2019task2vec}. The semantic embedding component captures fine-grained semantic relatedness, while the structural feature component complements it by enabling reliable task grouping based on structural correspondence, particularly in cases where semantic similarity alone is insufficient \cite{wang2024towards}. This embedding-level similarity assessment provides a principled foundation for cross-task knowledge transfer, in which tasks that are proximate in the task signature space can share agent weight configuration experience. 

\subsubsection{Meta-Learning Weight Predictor}

The meta-learning weight predictor leverages task signature patterns to associate the current task with previously trained tasks in the knowledge base. Through this weight predictor, a mapping function $\phi: \mathbf{s}_T \mapsto \hat{\mathbf{w}}$ is learned to directly output an agent weight vector $\hat{\mathbf{w}} = (w_1, \dots, w_N)$ from the task signature. This predicts the matching degree between this weight vector and the weight vectors in the knowledge base, thereby obtaining the best-performing weight vector for this reasoning task as derived in Section \ref{sec:3.1.1}. The prediction network adopts a two-layer MLP architecture \cite{rosenblatt1958perceptron}, and the output weights are constrained to satisfy predefined bounds and a simplex normalization, ensuring their validity. Training data is obtained by running SIAs on the training task set $\{T_k\}$ to get optimal weights $\mathbf{w}_{T_k}^*$. The loss function is defined as:

\begin{equation}
\mathcal{L}(\phi) = \sum_k \|\hat{\mathbf{w}}_{T_k} - \mathbf{w}_{T_k}^*\|_2^2 .
\end{equation}

This mean-squared error objective drives the predictor to learn a cross-task mapping from task characteristics to agent contribution patterns, enabling zero-shot prediction of task-adaptive weight configurations on unseen reasoning tasks.

\subsection{Weak-link Optimization}
\label{sec3.2}
Based on the ``weak-link'' optimization principle, WORC implements targeted compensation for identified weak agents. When using the most closely matched weight vector $\hat{\mathbf{w}}$ from the knowledge base to guide budget allocation, according to the theory in Section \ref{sec:3.1.1}, a lower numerical value indicates poorer performance of that agent.

With a total budget of $B$ additional reasoning opportunities, the allocation formula is:

\begin{equation}
m_i = \left\lfloor B \cdot \frac{\exp(\tau(1-w_i))}{\sum_i \exp(\tau(1-w_i))} \right\rceil,
\end{equation}
where $\tau$ controls the degree of bias toward weak agents, and $\lfloor \cdot \rceil$ represents rounding to the nearest integer. This strategy ensures that agents with lower weights receive more opportunities for repeat generation, thereby compensating for performance shortfalls. Additionally, the content from previous repeated generations serves as context for subsequent repeated generation, ensuring that repetition is not random generation but rather generation with experiential guidance.

\subsection{Illustrative Framework for Reasoning Optimization}
This section presents the AgentChain (AC) framework used to demonstrate WORC and summarizes the overall process.

\subsubsection{Structure of AgentChain}
\label{sec:chain}

AgentChain is conceptualized as a linear chain with self-loops, facilitating a clear presentation of the WORC method. Formally, it is represented as a graph \(G = (\mathcal{A}, E \cup S)\), where \(\mathcal{A}\) is a set of specialized agents, \(E\) represents directed edges encoding sequential information flow, and \(S\) represents self-loops that allow agents to repeatedly output information to compete for additional budget allocation.

The agent set \(\mathcal{A} = \{A_{DC}, A_{PU}, A_{SR}, A_{SP}\}\) comprises four core components. The Data Collection Agent \(A_{DC}\) collects relevant data based on the input problem \(P\) and prompt \(T1\), producing a knowledge set \(K_P = f_{DC}(P, T1)\) with functions covering target clarification, standardized data collection, challenge prediction, and quality verification. The Problem Understanding Agent \(A_{PU}\) integrates \(P\), \(K_P\), and prompt \(T2\) to produce a structured understanding \(R_P = f_{PU}(P, K_P, T2)\), focusing on mining relationships between known conditions and implicit information such as numerical or boundary constraints. The Step Reasoning Agent \(A_{SR}\) develops reasoning based on \(R_P\) and prompt \(T3\), generating a set of possible reasoning paths \(S_P = \{S_P^1, ..., S_P^m\}\) and selecting a path \(S_P^*\) that maximizes logical consistency and completeness. The Problem Solving Agent \(A_{SP}\) generates multiple candidate solutions \(\{Y_P^1, ..., Y_P^k\}\) based on the chosen path \(S_P^*\) and prompt \(T4\), then selects a final solution \(Y_P^*\) while recording solution steps and decision rationale.

Information flows along the directed edges \((A_i, A_j) \in E\) in a predefined sequential order. The self-loops in \(S\) enable repeated activation of agents to adjust resource allocation and support iterative reasoning, allowing the system to dynamically optimize performance based on budget allocation strategies.

\subsubsection{Other Design and Workflow Overview}
\label{sec:3.2.2}
Textual outputs are often challenging to evaluate directly, requiring WORC to be applied to multi-agent reasoning through bridging mechanisms. Therefore, in this work, we design the Eval Agent and the Vote Agent.

EvalAgent is used for the construction of the weight knowledge base in Section \ref{sec:3.1.1}, binding the iterative updates of SIAs with the actual output quality of each agent, providing a quantitative basis for dynamic adjustment of weights by scoring the single output of each agent adaptively. EvalAgent scores the output of agent $A_i$ in the current iteration $t$, resulting in quality score $sco_j^t$. These scores are normalized into a probability distribution, forming a guidance vector $\mathbf{G}^t$, to reflect the ideal weight distribution tendency in this iteration:
\begin{equation}
\mathbf{G}^t = \left( \frac{s_1^t}{\sum s_k^t}, \frac{s_2^t}{\sum s_k^t}, \ldots, \frac{s_N^t}{\sum s_k^t} \right)
\end{equation}
This guidance vector is integrated into the weight update formula, tightly binding the optimization process with specific output quality, allowing the weight vector to accurately capture the instantaneous performance contribution of each agent.

\begin{algorithm}
\caption{WORC: Weak-link Optimization for Reasoning Cooperation}
\label{alg:1}
\begin{algorithmic}[1]
\REQUIRE Multi-agent framework $\mathcal{F} = (\mathcal{A}, E)$ where $\mathcal{A} = \{A_1, A_2, \ldots, A_n\}$, Task $T$, Budget $B \in \mathbb{Z}^+$, trained meta-weight predictor $\phi$
\ENSURE Solution $Y_T^*$

\STATE \textbf{Weak Agent Localization Phase:}
\STATE $\mathcal{D} \leftarrow \{T_1, T_2, \ldots, T_k\}$ \COMMENT{Diverse reasoning task set}
\STATE Initialize knowledge base $\mathcal{K} \leftarrow \emptyset$
\FOR{each $T_i \in \mathcal{D}$}
    \STATE $\mathbf{w}_i^* \leftarrow \text{SIAsOptimize}(\mathcal{F}, T_i)$ \COMMENT{Optimal agent weights via SIAs}
    \STATE $\mathbf{s}_i \leftarrow \text{TaskSignature}(T_i)$
    \STATE $\mathcal{K} \leftarrow \mathcal{K} \cup \{(\mathbf{s}_i, \mathbf{w}_i^*)\}$
\ENDFOR
\STATE Train meta-weight predictor $\phi$ on $\mathcal{K}$

\STATE \textbf{weak-link Optimization Phase:}
\STATE $\mathbf{s}_T \leftarrow \text{TaskSignature}(T)$
\STATE $\hat{\mathbf{w}} \leftarrow \phi(\mathbf{s}_T)$
\STATE $Z \leftarrow \sum_{j=1}^n \exp(\tau(1-\hat{w}_j))$

\FOR{each agent $A_i \in \mathcal{A}$}
    \STATE $m_i \leftarrow \left\lfloor B \cdot \frac{\exp(\tau(1-\hat{w}_i))}{Z} \right\rceil$
\ENDFOR

\STATE \textbf{Execute Framework:}
\STATE $X_0 \leftarrow T$
\FOR{$i = 1$ to $n$}
    \STATE $O_i \leftarrow \{A_i(X_{i-1})^j : j \in \{1,2,...,m_i\}\}$
    \STATE $X_i \leftarrow \text{VoteAgent}(O_i)$
\ENDFOR
\STATE $Y_T^* \leftarrow X_n$

\RETURN $Y_T^*$
\end{algorithmic}
\end{algorithm}

Each agent executes a complete reasoning process in the sample set, generates responses using an LLM, and calculates a fitness score using the Eval agent to measure the alignment between the outputs and standard answers. This fitness score directly reflects the performance quality of the current weight configuration on the task. Subsequently, the solutions are ranked according to their fitness scores, and the weight vector demonstrating optimal performance is selected from the set of candidate solutions as $X_{curr}^t$. The iterations then proceed according to the SIAs configuration, with this ranking method indirectly reducing the differences in weight value outputs among different SIA. The final weights obtained, where each value represents the performance of the corresponding agent, with higher values indicating higher rankings.

VoteAgent is used for the selection of multiple output results from a single agent in a single reasoning task in Section \ref{sec3.2}. For each set of reasoning results output by an agent, the results in the set are scored and evaluated similar to EvalAgent, and the best and most likely correct result in the reasoning process is selected as the output to be passed to subsequent steps or as the final output. The scoring criteria of these two Agents rely on preset prompts to perform real-time quality assessment of intermediate results produced by agents in reasoning tasks, with evaluation dimensions strictly following the core requirements of the task's inherent logical rigor, semantic accuracy, and reasoning coherence, which are detailed in the supplementary materials. The overall workflow is summarized in Pseudocode \ref{alg:1}.

\begin{table*}[h!]
    \caption{Performance Comparison on Reasoning Benchmarks (\%). F1 scores are reported for HotpotQA and LongBench (tasks with partially correct answers), while exact match accuracy is used for other tasks with singular correct answers.}
    \label{tab:performance}
    \centering
    \small
    \begin{tabular}{l|ccccccc}
        \hline
        Method & MATH & GSM8K & BBH & MMLU-CF & HotpotQA & LongBench & Avg. \\
        \hline
        CoT & 78.3 & 90.9 & 78.3 & 69.6 & 67.2 & 57.6 & 73.6 \\
        CoT-SC ($n$=5) & 81.8 & 92.0 & 83.4 & 71.1 & 66.2 & 58.6 & 75.5 \\
        Self-Refine & 78.7 & 91.7 & 80.0 & 69.7 & 68.3 & 58.2 & 74.4 \\
        Analogical Prompting & 65.4 & 87.2 & 72.5 & 65.8 & 64.7 & 52.9 & 68.1 \\
        AFlow & 83.0 & 93.5 & 76.0 & 69.5 & 73.5 & 61.0 & 76.1 \\
        FoT ($n$=8) & 82.5 & 94.0 & 82.4 & 70.6 & 66.7 & 59.1 & 75.9 \\
        AoT & 83.6 & 95.0 & 86.0 & 70.9 & 80.6 & 68.5 & 80.8 \\
        AgentChain &81.0 $\pm$ 0.8 &92.4 $\pm$ 0.4 &81.1 $\pm$ 0.7 &70.3 $\pm$ 1.0 &75.1 $\pm$ 1.3 &64.3 $\pm$ 1.6 &77.4\\
        \hline
        \textbf{WORC+AC (Ours)} & $\mathbf{87.0 \pm 0.5}$ & $\mathbf{95.9 \pm 0.1}$ & $\mathbf{86.9 \pm 0.2}$ & $\mathbf{71.7 \pm 0.4}$ & $\mathbf{83.2 \pm 0.3}$ & $\mathbf{68.4 \pm 0.4}$ & $\mathbf{82.2 \pm 0.4}$ \\
        \hline
    \end{tabular}
\end{table*}

\section{Experiments}

\subsection{Experimental Setup}
\label{sec:exp_setup}
\textbf{Datasets.} To evaluate WORC’s reasoning ability, we conduct experiments on six benchmark datasets including MATH~\cite{hendrycks2021measuring} for advanced mathematical reasoning, GSM8K~\cite{cobbe2021training} for grade-school mathematical word problems, BBH~\cite{suzgun2022challenging} for logical and algorithmic reasoning tasks, MMLU-CF~\cite{zhao2024mmlu} for commonsense and factual knowledge evaluation, HotpotQA~\cite{yang2018hotpotqa} for multi-hop question answering, and LongBench~\cite{bai2023longbench} for long-context reasoning scenarios. These datasets collectively cover a wide spectrum of reasoning tasks and provide a comprehensive evaluation testbed for the framework.

\textbf{Implementation Details.} As mentioned earlier, WORC functions as an optimization method for multi-agent reasoning. For demonstration and comparison, it is implemented within the AgentChain (AC) framework, where WORC denotes the optimized system and AC refers to the baseline framework without optimization. All agents and methods are powered by GPT-4o, an LLM with strong reasoning capabilities and high efficiency suitable for multi-agent collaboration. The framework implements three SIAs, namely Hippopotamus Optimization Algorithm \cite{amiri2024hippopotamus}, Particle Swarm Optimization \cite{kennedy1995particle}, and Grey Wolf Optimizer \cite{mirjalili2014grey}, which are randomly selected as representative examples for demonstration rather than tailored to specific task characteristics. The architecture is built on Langchain using official API calls.

These simplified SIAs are distilled into three primary factors (Complete details shown in supplementary materials). In the main results, the HO is set with parameters as $\alpha$ (individual update) = 0.9, $\beta$ (group interaction) = 0.1, $\gamma$ (exploration) = 0.05, with a population size of 5, 3 iterations, and 10 samples for few-shot learning on each dataset. The recommended parameter ranges maintain $\alpha \geq \beta$ and $\alpha \geq \gamma$, with other parameters following the original algorithm. The PSO is set $w$ (inertia weight) = 0.8, $c_1$ (cognitive learning factor) = 1.5, $c_2$ (social learning factor) = 1.0 and maximum velocity = 0.2. The GWO's maximum iterations is set to 100.

\textbf{Baselines.} We compare WORC with several representative reasoning methods. Chain-of-Thought (CoT)~\cite{wei2022chain} guides LLMs to solve problems by generating intermediate reasoning steps. CoT-SC enhances CoT by sampling multiple reasoning paths ($n=5$) and selecting the most consistent answer. Self-Refine~\cite{madaan2023self} iteratively refines outputs through multi-pass generation. Analogical Prompting~\cite{yasunaga2023large} leverages analogical reasoning patterns in prompts. AFlow~\cite{zhang2024aflow} is a feedback-driven reasoning framework that dynamically adjusts its reasoning process. FoT applies a fine-tuned CoT-style model and selects answers via voting over 8 sampled outputs \cite{bi2025forest}. AoT \cite{teng2025atom} enhances LLM performance by decomposing complex reasoning into Markov thought atomic units during inference.

\begin{table*}[h]
\centering
\caption{Performance comparison of different optimization methods across three MAS architectures.}
\label{tab:mas_opt}
\begin{tabular}{l l c c c c}
\toprule
\textbf{MAS Architecture} & \textbf{Optimization Method} & \textbf{MATH Acc (\%)} & \textbf{BBH Acc (\%)} & \textbf{HotpotQA F1 (\%)} & \textbf{Average (\%)} \\
\midrule
\multirow{4}{*}{MetaGPT}
& None & 81.9 & 82.7 & 78.0 & 81.0 \\
& AFlow & 83.2 & 83.9 & 79.6 & 81.8 \\
& Majority Voting & 84.0 & 84.6 & 80.4 & 82.7 \\
& \textbf{WORC} & \textbf{86.5} & \textbf{86.8} & \textbf{80.8} & \textbf{85.0} \\
\midrule
\multirow{4}{*}{HIMA}
& None & 83.5 & 83.8 & 79.0 & 82.1 \\
& AFlow & 84.2 & 84.4 & 79.6 & 82.7 \\
& Majority Voting & 85.1 & 85.3 & 80.6 & 83.7 \\
& \textbf{WORC} & \textbf{87.2} & \textbf{87.0} & \textbf{82.0} & \textbf{85.4} \\
\midrule
\multirow{4}{*}{MAS$^2$}
& None & 84.5 & 84.8 & 80.5 & 83.3 \\
& AFlow & 85.1 & 85.4 & 81.0 & 83.8 \\
& Majority Voting & 86.0 & 86.2 & 81.9 & 84.7 \\
& \textbf{WORC}& \textbf{88.0} & \textbf{88.2} & \textbf{82.7} & \textbf{86.3} \\
\midrule
\multirow{4}{*}{AgentChain}
& None & 81.0 & 81.1 & 75.1 & 79.1 \\
& AFlow & 83.5 & 83.3 & 77.2 & 81.3 \\
& Majority Voting & 85.0 & 84.6 & 79.0 & 82.9 \\
& \textbf{WORC} & \textbf{87.0} & \textbf{86.9} & \textbf{83.2} & \textbf{85.7} \\
\bottomrule
\end{tabular}
\end{table*}

In addition to inference-level reasoning baselines, WORC is evaluated as a framework-level optimization strategy within multiple multi-agent system (MAS) architectures. Specifically, WORC is integrated into four representative MAS frameworks, including MetaGPT \cite{hong2023metagpt},HIMA \cite{ahn2025society}, MAS$^2$ \cite{wang2025mas}, and AgentChain, and compared against commonly used optimization strategies such as Majority Voting \cite{choi2025debate} and AFlow \cite{zhang2024aflow}. For each architecture, we report the performance of the original system without optimization, as well as its variants enhanced by different optimization methods. To ensure fair comparison across reasoning tasks, the core architectural designs of these MAS frameworks are preserved, while their task-specific prompts are unified into an AgentChain-style reasoning format for consistent task decomposition and inter-agent interaction.

\subsection{Performance Evaluation}
\label{sec:performance}     
\textbf{Main Results.} Table~\ref{tab:performance} shows that our WORC method achieves competitive performance across all benchmarks, with an average accuracy of 82.22\% $\pm$ 0.4\%. This represents a significant improvement over previous methods, outperforming FoT by 6.3\% and AFlow by 6.1\%. The standard deviations ($\pm$0.1-0.5\%) indicate strong consistency across different SIA implementations. WORC demonstrates particularly impressive gains on complex reasoning tasks, achieving 83.2\% on HotpotQA and 68.4\% on LongBench, validating the effectiveness of our optimization method.

\textbf{Generalization Comparison.}
Table~\ref{tab:mas_opt} further reports the performance of WORC when deployed as an optimization module across multiple MAS architectures. Under all architectural settings, the integration of WORC leads to consistent performance improvements over the original systems without optimization, with average gains of 4.0\%, 3.3\%, 3.0\%, and 6.6\% on MetaGPT, HIMA, MAS$^2$, and AgentChain, respectively. Compared with alternative optimization strategies such as Majority Voting and AFlow, WORC exhibits improved reasoning accuracy across all evaluated benchmarks. These observations suggest that the proposed optimization mechanism provides a task-adaptive enhancement effect that generalizes across multi-agent frameworks.

\begin{table*}[t]
\centering
\caption{Ablation study on task signature components. Performance is reported in accuracy (\%).}
\label{tab:signature_ablation}
\begin{tabular}{l c c c c c c c}
\toprule
\textbf{Configuration} & \textbf{MATH} & \textbf{GSM8K} & \textbf{BBH} & \textbf{MMLU-CF} & \textbf{HotpotQA} & \textbf{LongBench} & \textbf{Avg $\pm$ Std} \\
\midrule
Semantic Embedding Only 
& 85.4 $\pm$ 0.6 
& 95.3 $\pm$ 0.2 
& 84.1 $\pm$ 0.4 
& 69.5 $\pm$ 0.6 
& 79.0 $\pm$ 0.5 
& 66.1 $\pm$ 0.6 
& 79.9 $\pm$ 0.5 \\

Statistical Features Only 
& 84.0 $\pm$ 0.7 
& 94.2 $\pm$ 0.3 
& 84.0 $\pm$ 0.4 
& 67.6 $\pm$ 0.5 
& 78.8 $\pm$ 0.4 
& 65.7 $\pm$ 0.5 
& 80.0 $\pm$ 0.5 \\

\textbf{Full Task Signature} 
& \textbf{87.0 $\pm$ 0.5}
& \textbf{95.9 $\pm$ 0.1}
& \textbf{86.9 $\pm$ 0.2}
& \textbf{71.7 $\pm$ 0.4}
& \textbf{83.2 $\pm$ 0.3}
& \textbf{68.4 $\pm$ 0.4}
& \textbf{82.2 $\pm$ 0.3} \\
\bottomrule
\end{tabular}
\end{table*}

\begin{table*}[t]
\centering
\caption{Comparison of different budget allocation strategies for weak-link compensation. Performance is reported in accuracy (\%).}
\label{tab:allocation_strategy}
\begin{tabular}{l c c c c c c c}
\toprule
\textbf{Allocation Strategy} & \textbf{MATH} & \textbf{GSM8K} & \textbf{BBH} & \textbf{MMLU-CF} & \textbf{HotpotQA} & \textbf{LongBench} & \textbf{Average (\%)} \\
\midrule
Uniform Allocation 
& 84.1 & 94.3 & 84.2 & 70.4 & 80.0 & 66.8 & 80.0 \\

Predefined Rule-Based Allocation 
& 85.3 & 95.1 & 85.4 & 71.0 & 81.0 & 67.5 & 80.9 \\

\textbf{WORC Allocation} 
& \textbf{87.0} & \textbf{95.9} & \textbf{86.9} & \textbf{71.7} & \textbf{83.2} & \textbf{68.4} & \textbf{82.2} \\
\bottomrule
\end{tabular}
\end{table*}

\textbf{Comparative Results of Different LLMs.} 
We compare the performance of the WORC method driven by three different LLMs, including DeepSeek-V3\cite{liu2024deepseek}, Qwen-Turbo, and GPT-4.1-nano. We make comparisons on reasoning benchmarks on the same scale as in Table 1. Among these, the DeepSeek-V3 powered implementation demonstrates superior performance, achieving 89.3 in MATH and 98.2 on GSM8K, significantly outperforming both Qwen-Turbo and GPT-4.1-nano on all benchmarks with leads typically ranging from 1.5 to 5.8 percentage points. The latter two models perform similarly overall but with mixed results. GPT-4.1-nano slightly outperformed on BBH and MMLU-CF, while Qwen-Turbo showed minimal differences on the same tasks.

\begin{table}[h]
\label{tab:model_comparison}
\caption{Performance Comparison across Different LLMs. Three different LLMs driving the WORC method were compared on six reasoning benchmarks (\%), with the configurations and evaluation metrics identical to those in Table~\ref{tab:performance}.} 
\centering
\small
\begin{tabular}{lccc} 
\hline
\textbf{Dataset} & DeepSeek V3 & Qwen-Turbo & GPT-4.1 nano \\ 
\hline
MATH        & 89.3 & 84.9 & 85.3 \\
GSM8K       & 98.2 & 95.3 & 95.6 \\
BBH         & 88.7 & 84.2 & 85.7 \\
MMLU-CF     & 75.3 & 69.5 & 72.3 \\
HotpotQA    & 85.0 & 81.2 & 83.5 \\
LongBench   & 75.5 & 70.8 & 70.1 \\
\hline
\end{tabular}
\end{table}

\subsection{Ablation Study}
In this section, we conduct systematic ablation studies to analyze the contribution of each component in WORC. In particular, the impact of the SIAs is evaluated independently to isolate its role.

\textbf{Task Signature Component Analysis.} 
Table~\ref{tab:signature_ablation} reports the ablation results on the task signature design. Removing either semantic embeddings or statistical features leads to consistent performance degradation across all benchmarks. The full task signature achieves the highest average accuracy (82.2\% $\pm$ 0.3), outperforming the semantic-only and statistical-only variants by 2.3 and 2.2 percentage points, respectively. Performance gaps are more pronounced on structurally complex tasks such as MMLU-CF and HotpotQA, where the full representation improves over single-component variants by more than 2 points. These results indicate that semantic representations serve as the primary basis for weight prediction on unseen tasks, while structural statistics provide auxiliary task-level signals that further refine the predicted weight configuration.

\textbf{Weak-Link Compensation Strategy.} 
Table~\ref{tab:allocation_strategy} compares different budget allocation strategies for weak-link compensation to validate the effectiveness of dynamic resource-aware reasoning allocation. For the AC baseline, the uniform allocation requires one additional reasoning step for each agent, while the predefined rule-based allocation assigns two additional reasoning steps to each of the two lowest-ranked agents. Uniform allocation yields an average accuracy of 80.0\%, while rule-based allocation improves performance to 80.9\%. In contrast, the WORC allocation strategy achieves 82.2\%, consistently outperforming alternative strategies across all tasks. The relative gains are more evident on long-context and multi-hop reasoning benchmarks such as HotpotQA and LongBench, where adaptive allocation improves performance by 3.2 and 1.6 percentage points over uniform allocation, respectively. This pattern suggests that weight-aware adaptive budget redistribution better captures task-specific reasoning difficulty than static or predefined strategies.

\textbf{Cross-Task Generalization.} 
Table~\ref{tab:cross_task} evaluates the cross-task generalization capability of the learned weight predictor. When trained on one dataset and evaluated on another, WORC preserves competitive performance with moderate variance. For instance, training on GSM8K and testing on MATH yields 86.3\% accuracy, while the reverse setting attains 95.6\% on GSM8K. Consistent trends are observed across other task pairs, including MMLU-CF and BBH, as well as HotpotQA and LongBench. Although cross-domain transfer results in slight performance degradation relative to in-domain training, the overall performance remains stable, suggesting that the proposed task signature and meta-learned weight predictor capture transferable reasoning characteristics beyond dataset-specific patterns.

\begin{table}[h]
\centering
\caption{Cross-task generalization performance of WORC.}
\label{tab:cross_task}
\begin{tabular}{l c}
\toprule
\textbf{Train $\rightarrow$ Test} & \textbf{Accuracy (\%)} \\
\midrule
MATH $\rightarrow$ GSM8K & 95.6 $\pm$ 0.2 \\
GSM8K $\rightarrow$ MATH & 86.3 $\pm$ 0.6 \\
MMLU-CF $\rightarrow$ BBH & 86.0 $\pm$ 0.4 \\
BBH $\rightarrow$ MMLU-CF & 71.3 $\pm$ 0.5 \\
HotpotQA $\rightarrow$ LongBench & 67.4 $\pm$ 0.5 \\
LongBench $\rightarrow$ HotpotQA & 81.0 $\pm$ 0.4 \\
\bottomrule
\end{tabular}
\end{table}

\begin{table*}[t]
\centering
\caption{Performance comparison under different optimization strategies. \textbf{Variation} denotes the averaged performance fluctuation across tasks (lower is better).}
\label{tab:optimizer}
\setlength{\tabcolsep}{6pt}
\begin{tabular}{lccccccc}
\toprule
Method & MATH & GSM8K & BBH & MMLU-CF & HotpotQA & LongBench & Variation $\downarrow$ \\
\midrule
AgentChain (AC) & 81.0 $\pm$ 0.8 & 92.4 $\pm$ 0.4 & 81.1 $\pm$ 0.7 & 70.3 $\pm$ 1.0 & 75.1 $\pm$ 1.3 & 64.3 $\pm$ 1.6 & 0.97 \\
WORC+AC (GWO) & 86.6 $\pm$ 0.5 & 95.8 $\pm$ 0.2 & 86.4 $\pm$ 0.3 & 71.6 $\pm$ 0.4 & 81.5 $\pm$ 0.4 & 68.0 $\pm$ 0.4 & 0.37 \\
WORC+AC (PSO) & 86.6 $\pm$ 0.6 & 96.0 $\pm$ 0.2 & 86.3 $\pm$ 0.3 & 71.8 $\pm$ 0.5 & 81.5 $\pm$ 0.4 & 67.9 $\pm$ 0.5 & 0.42 \\
WORC+AC (HO) & 87.0 $\pm$ 0.5 & 95.9 $\pm$ 0.1 & 86.9 $\pm$ 0.2 & 71.7 $\pm$ 0.4 & 83.2 $\pm$ 0.3 & 68.4 $\pm$ 0.4 & 0.32 \\
\bottomrule
\end{tabular}
\end{table*}

\begin{figure*}[h]
  \centering
  \includegraphics[width=\linewidth]{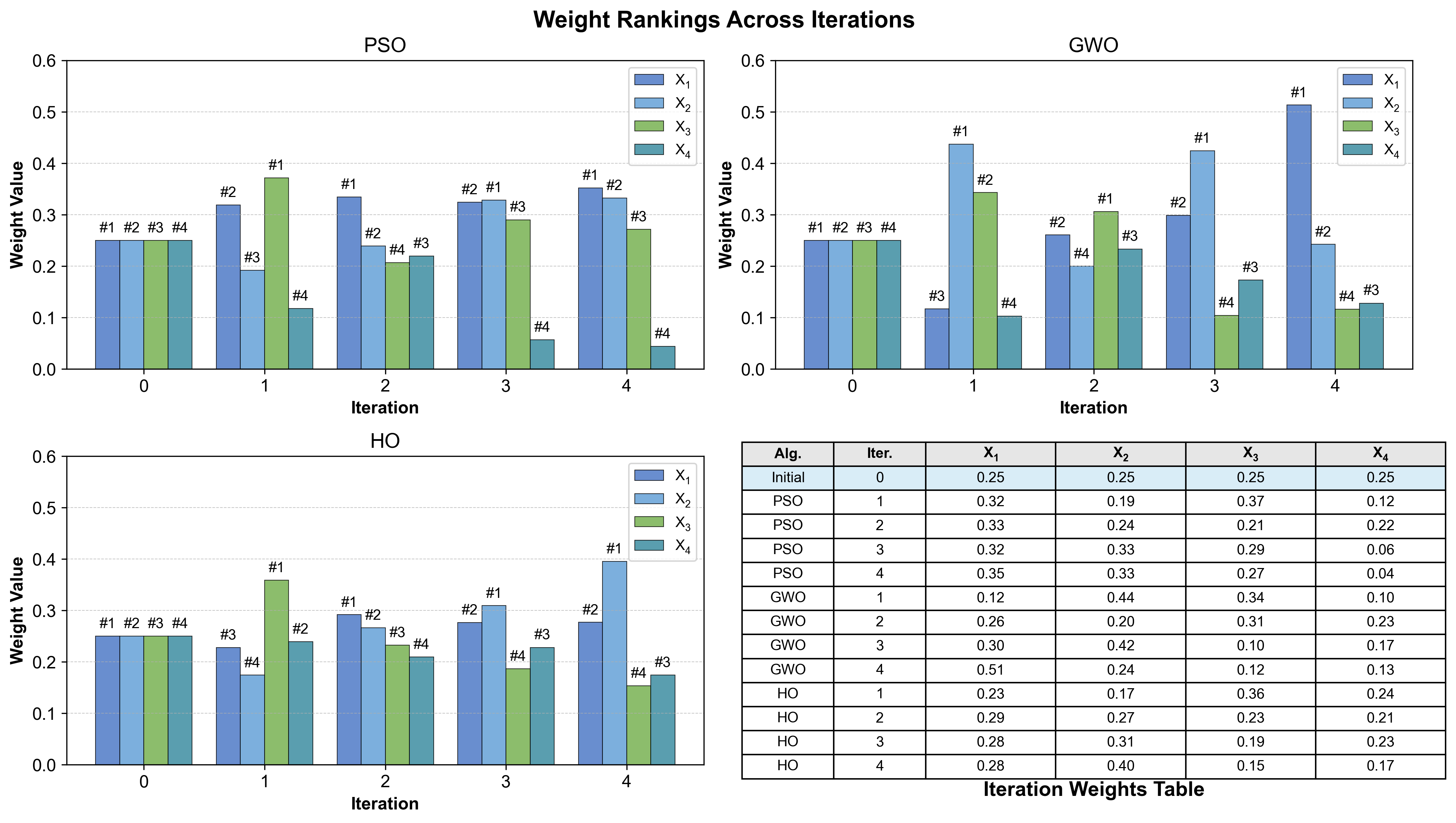}
  \caption{Agent weight evolution across iterations under three SIAs (PSO, GWO, and HO). The labels \#1-\#4 denote the relative ranking of the four agents at each iteration, from the highest-weighted to the lowest-weighted agent. Although the absolute weight values and convergence trajectories vary across SIAs, the ranking patterns remain largely consistent, indicating that weak-agent identification in WORC mainly depends on stable relative rankings rather than exact weight magnitudes.}
  \label{fig:rank}
\end{figure*}

\subsection{Different SIAs Analysis}
SIAs are used to estimate agent effectiveness and generate task-specific weight vectors for coordination.  Although different SIAs may produce variations in weight magnitude and convergence dynamics, the downstream budget allocation is determined through a smooth normalization mechanism that emphasizes relatively weaker agents.  As a result, moderate numerical differences in weight estimation translate into only minor changes in the final allocation ratios. Empirically, this leads to comparable optimization trends and similar performance across SIA variants, suggesting that the effectiveness of WORC is primarily driven by the weak-link compensation principle rather than the precise choice of swarm optimization algorithm. The observed stability indicates that the framework maintains consistent coordination behavior under different weight estimation processes.

We compare HO, GWO, and PSO within the AC framework, with the corresponding weight evolution and rank patterns shown in Figure~\ref{fig:rank}. Here, $X_1$--$X_4$ denote the four agent-specific weights in AC, corresponding to the four agents, respectively. They characterize the relative contribution assigned to each agent during iterative optimization. Across different SIAs, the identified weak agents remain largely consistent, although moderate variations in weight magnitude are observed. As reflected by the accuracy fluctuations reported in Table~\ref{tab:performance}, pairing WORC with different SIAs introduces dataset-dependent performance variance. While these differences indicate that the choice of SIA influences the learned weight configuration to some extent, the resulting allocation patterns and optimization trends remain generally stable. It is worth noting that, given the diversity of swarm-based optimization algorithms, achieving complete invariance across SIAs is difficult in practice. Nonetheless, the observed consistency suggests that WORC retains its effectiveness under reasonable variations in weight estimation, highlighting both the feasibility and the inherent limitations of algorithm-agnostic coordination.

\textbf{Impact of SIAs.} Table~\ref{tab:optimizer} summarizes the performance of WORC when combined with different SIAs on the AC framework. Consistent improvements over the baseline AC are observed across all six benchmarks under all optimization strategies. In particular, WORC yields clear accuracy gains on mathematical reasoning tasks such as MATH and GSM8K, as well as on multi-step reasoning and knowledge-intensive benchmarks including BBH, MMLU-CF, and HotpotQA. Performance improvements are also evident on long-context tasks such as LongBench. Moreover, all WORC variants demonstrate substantially reduced performance fluctuation compared to AC, indicating enhanced stability in multi-agent coordination. These findings suggest that the proposed optimization mechanism maintains consistent effectiveness across different SIA implementations while improving both reasoning accuracy and coordination robustness. Together with the rank consistency observed in Figure~\ref{fig:rank}, these results indicate that the effectiveness of WORC is primarily attributed to stable weak-agent identification and compensation, rather than dependence on any single SIA implementation.

\begin{figure}[H]
  \centering
  \includegraphics[width=0.95\columnwidth]{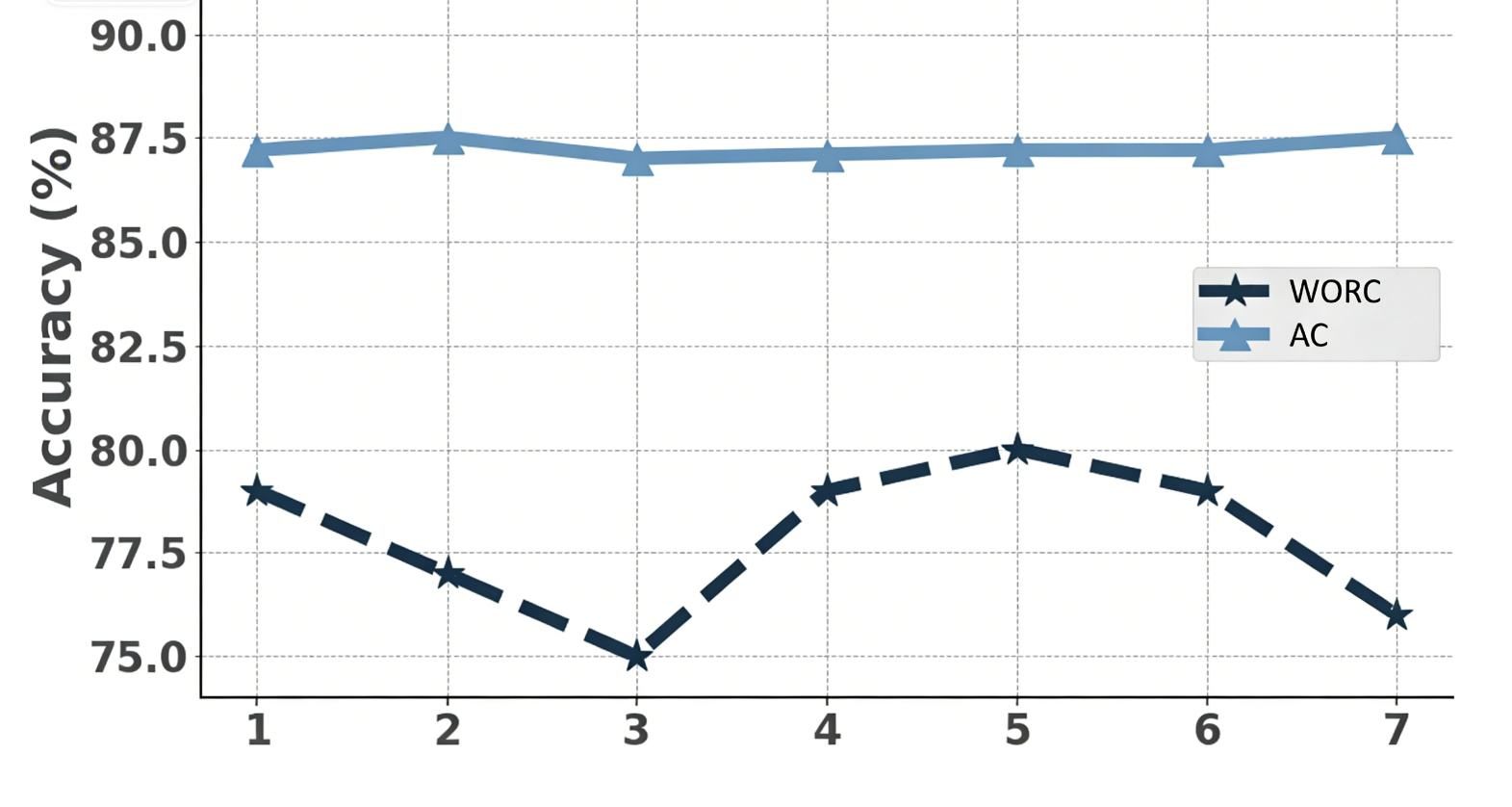}
  \caption{Compares Accuracy between AC and WORC method  (same configuration as main result experiment) across multiple trials on the MATH dataset. WORC consistently achieves high and stable accuracy, while AC shows greater variability, indicating lower stability.}
  \label{fig:example5}
\end{figure}

\textbf{Stability Analysis.} As shown in Figure~\ref{fig:example5}, WORC demonstrates more stable performance across multiple runs compared to AC, which the accuracy values represent the average across three SIAs. Specifically, AC exhibits accuracy fluctuations ranging from 75\% to 80\%, whereas WORC maintains a tighter range of 87\% to 87.5\%. This reduced variance suggests that WORC's training process is more robust. The instability observed in AC reflects a common limitation of multi-agent frameworks, which may compromise model reliability in practical applications, particularly for tasks requiring complex reasoning. Consequently, WORC offers enhanced stability, providing more dependable performance.

\subsection{Agreement Between EvalAgent and Human Experts}
To verify the objectivity and reliability of the EvalAgent in the WORC framework when assessing the quality of intermediate and final outputs produced by agents, we conduct a human-aligned evaluation study. The goal is to examine whether the score-based guidance signals derived from EvalAgent are consistent with human judgments. We select the GSM8K and HotpotQA datasets for comparative analysis, which represent numerical reasoning and semantically complex multi-hop question answering tasks, respectively. From the reasoning trajectories generated by the AC framework on these two datasets, we randomly sample 100 reasoning states from each dataset that are produced by different agents. These states include both partial solutions and final answers, resulting in a total of 200 evaluation samples.

We use a discrete 1-to-5 rating scale based on logical rigor, semantic correctness, and reasoning coherence. A score of 1 indicates severe logical errors and low-quality responses, whereas a score of 5 indicates logically sound and fully correct reasoning. EvalAgent first evaluates the 200 samples and produces corresponding quality scores. Subsequently, human experts independently assess the same samples under a blind evaluation setting in which the scores assigned by EvalAgent are not disclosed. The same rating scale is used to ensure consistency between the two evaluation processes. To quantify the level of agreement between EvalAgent and human experts, we adopt the Quadratic Weighted Cohen's Kappa coefficient $\kappa_w$ \cite{doewes2023evaluating}. This quadratic weighting scheme penalizes large disagreements more strongly, so that discrepancies such as scores of 1 and 5 receive substantially higher penalties than minor differences such as scores of 4 and 5.

\begin{table}[h]
\centering
\caption{Inter-rater reliability between EvalAgent and human experts. Following the standard interpretation of weighted Cohen's kappa, values of $0.61 \le \kappa_w \le 0.80$ indicate substantial agreement.}
\begin{tabular}{lccc}
\toprule
Dataset & Sample Size & $\kappa_w$ & Agreement Level \\
\midrule
GSM8K & 100 & 0.78 & Substantial \\
HotpotQA & 100 & 0.72 & Substantial \\
\bottomrule
\end{tabular}
\label{tab:evalagent_agreement}
\end{table}

\begin{table*}[t]
\centering
\caption{Efficiency and cost analysis of WORC compared with the baseline AgentChain (AC). Acc. denotes accuracy (\%), and pp denotes percentage points.}
\label{tab:cost_analysis}
\begin{tabular}{l c c c c c}
\toprule
\textbf{Dataset} & \textbf{AC Cost} & \textbf{WORC+AC Cost} & \textbf{AC Acc. (\%)} & \textbf{WORC+AC Acc. (\%)} & \textbf{Acc. Gain (pp)} \\
\midrule
MATH      & 1.20 & 2.30 & 83.9 & 86.7 & 2.8 \\
GSM8K     & 2.00 & 3.36 & 95.1 & 95.9 & 0.8 \\
BBH       & 1.60 & 2.76 & 84.2 & 86.5 & 2.3 \\
MMLU-CF   & 1.70 & 2.85 & 70.3 & 71.7 & 1.4 \\
HotpotQA  & 3.20 & 5.75 & 79.4 & 81.6 & 2.2 \\
LongBench & 18.0 & 33.0 & 65.3 & 68.0 & 2.7 \\
\midrule
Average   & 4.62 & 8.34 & 79.7 & 81.7 & 2.03 \\
\bottomrule
\end{tabular}
\end{table*}

As shown in Table \ref{tab:evalagent_agreement}, the EvalAgent in the WORC framework demonstrates stable and reliable evaluation performance. On the GSM8K task, the Quadratic Weighted Cohen's Kappa coefficient between EvalAgent and human experts reaches $\kappa_w = 0.78$, indicating a high level of agreement. This result suggests that EvalAgent can accurately identify high-quality intermediate reasoning steps and produce assessments that are highly consistent with human judgment. Furthermore, on the HotpotQA multi-hop question answering task, where textual expressions are highly diverse and evaluation can be more susceptible to potential length or formatting biases in large language models, EvalAgent still achieves $\kappa_w = 0.72$. This result indicates that EvalAgent remains robust under more complex semantic conditions. It effectively captures the true marginal contributions of different agents during the reasoning process and provides a reliable quantitative basis for the weight values derived from the SIA-based optimization procedure.


\subsection{Resource and Cost Analysis}
Table~\ref{tab:cost_analysis} summarizes the resource and performance trade-off of WORC relative to AC. Although WORC introduces additional test-time computation through adaptive weak-link compensation, it delivers consistent accuracy gains across all datasets, with an average improvement of 2.03 pp. The gains are especially clear on structurally complex tasks such as MATH and BBH, where WORC improves accuracy by 2.8 pp and 2.3 pp, respectively. These results suggest that the additional computation supports more reliable and higher-quality multi-agent reasoning. Notably, positive accuracy gains are maintained even on higher-cost benchmarks such as LongBench, indicating that the benefit of weak-link compensation is not limited to relatively inexpensive settings.

\section{Conclusion}
This paper presents WORC, a multi-agent reasoning optimization method grounded in a weak-link optimization perspective. By integrating SIAs with task signature construction, WORC enables cross-task weight generalization through meta-learning predictors. Under the guidance of these predicted weights, additional reasoning resources are allocated to weak agents, producing more consistent reasoning outcomes across tasks. Experimental results show that WORC achieves consistent gains in reasoning accuracy, stability, and cross-architecture generalization.  These findings indicate that weak-link compensation provides an effective and generalizable paradigm for improving multi-agent reasoning systems. Future work will focus on reducing computational overhead, improving online adaptability, and extending WORC to larger-scale and more heterogeneous multi-agent environments.





%
\bibliographystyle{IEEEtran}
\bibliography{IEEEabrv,IEEEexample}

\end{document}